\documentclass[10pt, conference, compsocconf]{IEEEtran}
\usepackage{multirow}
\usepackage{amssymb}

\PassOptionsToPackage{numbers, compress}{natbib}
\usepackage{hyperref}
\hypersetup{
    colorlinks=true,
    linkcolor=blue,
    filecolor=magenta,      
    urlcolor=cyan,
    pdftitle={Overleaf Example},
    pdfpagemode=FullScreen,
    }

\ifCLASSINFOpdf
\usepackage[pdftex]{graphicx}
\else
\fi
\usepackage[cmex10]{amsmath}
\usepackage{url}

\begin{document}
%
\title{LeYOLO, New Embedded Architecture for Object Detection}

\author{
    \IEEEauthorblockN{
        \textbf{Lilian Hollard}, Lucas Mohimont, Luiz Angelo Steffenel
    } \\
    \IEEEauthorblockA{
        Université de Reims Champagne-Ardenne, \\CEA, LRC DIGIT,\\ LICIIS, Reims, France \\
        Email: lilian.hollard, lucas.mohimont, luiz-angelo.steffenel \\
        @univ-reims.fr
    }
    \and
    \IEEEauthorblockN{Nathalie Gaveau} \\
    \IEEEauthorblockA{
        Université de Reims Champagne-Ardenne, \\INRAE, RIBP USC 1488, \\Reims, France \\
        Email: nathalie.gaveau@univ-reims.fr
    }
    
}


%


\maketitle

\begin{abstract}
Efficient computation in deep neural networks is crucial for real-time object detection. However, recent advancements primarily result from improved high-performing hardware rather than improving parameters and FLOP efficiency. This is especially evident in the latest YOLO architectures, where speed is prioritized over lightweight design. As a result, object detection models optimized for low-resource environments like microcontrollers have received less attention. For devices with limited computing power, existing solutions primarily rely on SSDLite or combinations of low-parameter classifiers, creating a noticeable gap between YOLO-like architectures and truly efficient lightweight detectors. This raises a key question: \textit{Can a model optimized for parameter and FLOP efficiency achieve accuracy levels comparable to mainstream YOLO models?} To address this, we introduce two key contributions to object detection models using MSCOCO as a base validation set. First, we propose \textit{LeNeck}, a general-purpose detection framework that maintains inference speed comparable to SSDLite while significantly improving accuracy and reducing parameter count. Second, we present \textit{LeYOLO}, an efficient object detection model designed to enhance computational efficiency in YOLO-based architectures. LeYOLO effectively bridges the gap between SSDLite-based detectors and YOLO models, offering high accuracy in a model as compact as MobileNets.  Both contributions are particularly well-suited for mobile, embedded, and ultra-low-power devices, including microcontrollers, where computational efficiency is critical.

Code : \url{https://github.com/LilianHollard/LeYOLO}.

\end{abstract}


%
\IEEEpeerreviewmaketitle

\section{Introduction}

Efficient computation, real-time processing, and low-latency execution are essential for AI-powered edge devices, including autonomous drones, surveillance systems, smart agriculture, and intelligent cameras. While cloud computing offers an alternative for running powerful models, it has drawbacks such as latency, bandwidth constraints, and security risks \cite{buyya_cloud_2009, zhou_edge_2019, wang_deep_2020}. In practical applications of object detection, deep learning advancements have largely focused on optimizing speed for high-end GPUs, often at the expense of efficiency on low-power hardware.

Initially introduced by Redmon et al.\cite{redmon_you_2016}, YOLO models are known for their inference speed in object detection. These models have seen significant architectural improvements in recent years, taking advantage of modern computing power.

Despite their inherent speed, there has been a noticeable shift in the development of YOLO models in recent years. With rapid advancements in GPU capabilities and new hardware innovations, the focus has shifted from lightweight models to those prioritizing inference speed \cite{jocher_ultralyticsyolov5_2022, li_yolov6_2022, wang_yolov7_2023, jocher_ultralytics_2023, wang_yolov9_2024}. Consequently, YOLO models have become significantly faster despite increased parameters and FLOP\footnote{We describe floating point operations as \textbf{FLOP}, defining all the number of arithmetical operations the neural network requires to perform inference. \\ In our paper, 1 FLOP is roughly 2 MADD or 2 MACC. Thus, the variation in benchmarks such as MobileNet differs from their original paper.}.

Our work highlights that despite their impressive speed on GPUs, YOLO models struggle with low-end hardware such as microcontrollers and embedded microcomputers. For instance, on STMicroelectronics microcontrollers—widely used in robotics and IoT applications—modern YOLO models take over a second per inference on the most powerful chips (Section \ref{chap:exp}, Table \ref{tab:speed}), making them impractical for real-time applications. On less powerful microcontrollers, further improvements are needed to reduce inference time, a challenge we address in this study. These constraints pose a critical challenge for industries relying on low-power AI, where power efficiency, small model size, and optimized resource usage are essential.

In classification tasks, research into optimizing parameter counts and computational costs has produced noteworthy models like MobileNets \cite{howard_mobilenets_2017, sandler_mobilenetv2_2018, howard_searching_2019} and EfficientNets \cite{tan_efficientnet_2019, tan_efficientnetv2_2021}. While these models are remarkable, they are primarily recognized for their exceptional classification abilities rather than object detection. Research has mainly focused on lightweight classifiers, often paired with an object detection addon like SSDLite \cite{liu_ssd_2016,sandler_mobilenetv2_2018}. While low-parameter classifiers combined with SSDLite offer better speed on microcontrollers, their accuracy falls short compared to YOLO.

Our research has identified a crucial gap: \textit{there is limited focus on optimizing object detection architectures that balance parameter efficiency and computational cost while maintaining YOLO-level accuracy.} This gap forces developers to choose between high-performance but computationally expensive YOLO models and low-power alternatives like SSDLite, which sacrifice accuracy for speed. Our work aims to bridge this divide by introducing more efficient object detection models tailored for edge AI applications.

\begin{figure}
    \centering
    \includegraphics[width=1\linewidth]{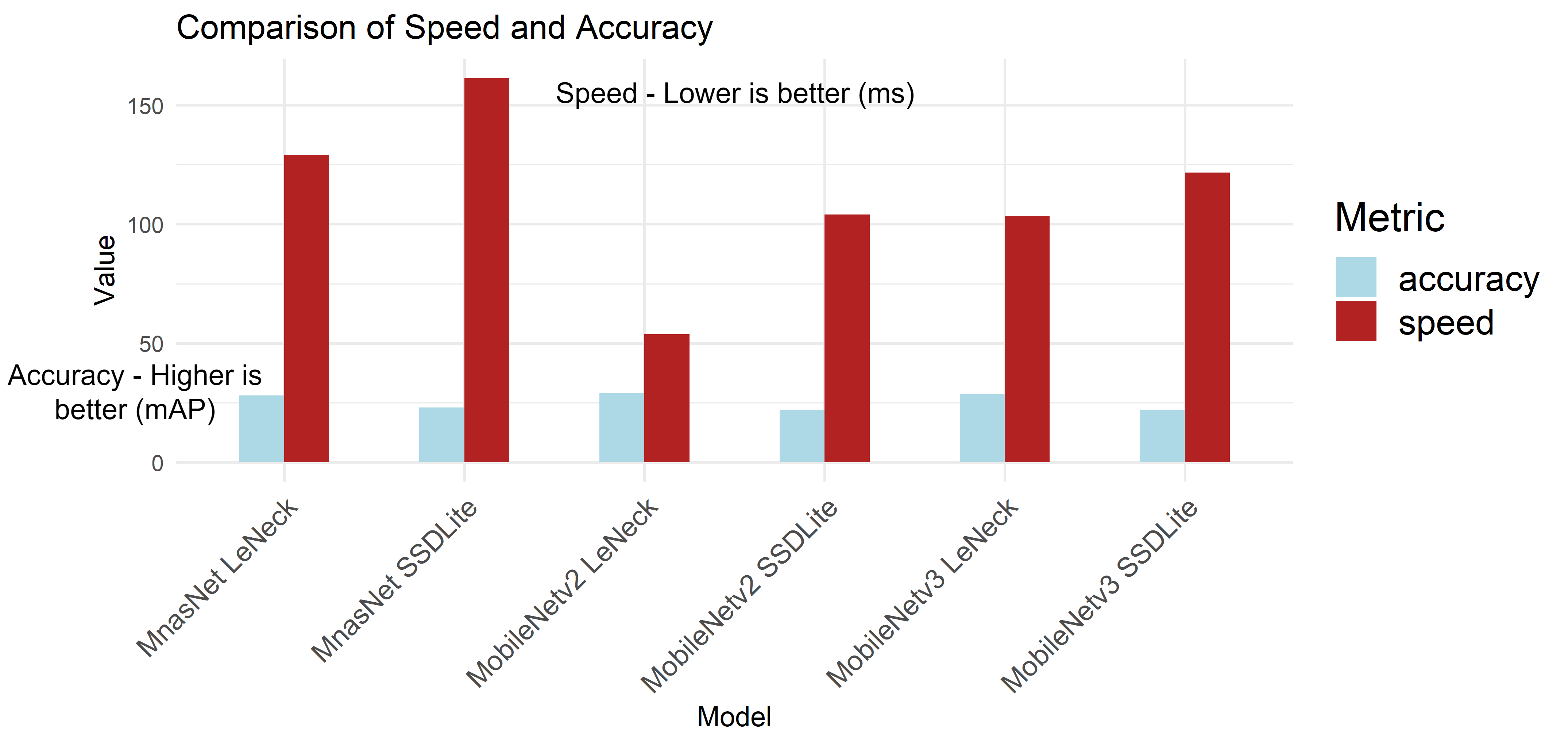}
    \caption{Speed and accuracy differences between SSDLite and LeNeck on STM32N6570-DK.}
    \label{fig:ssdlite}
\end{figure}

\begin{figure}
    \centering
    \includegraphics[width=1\linewidth]{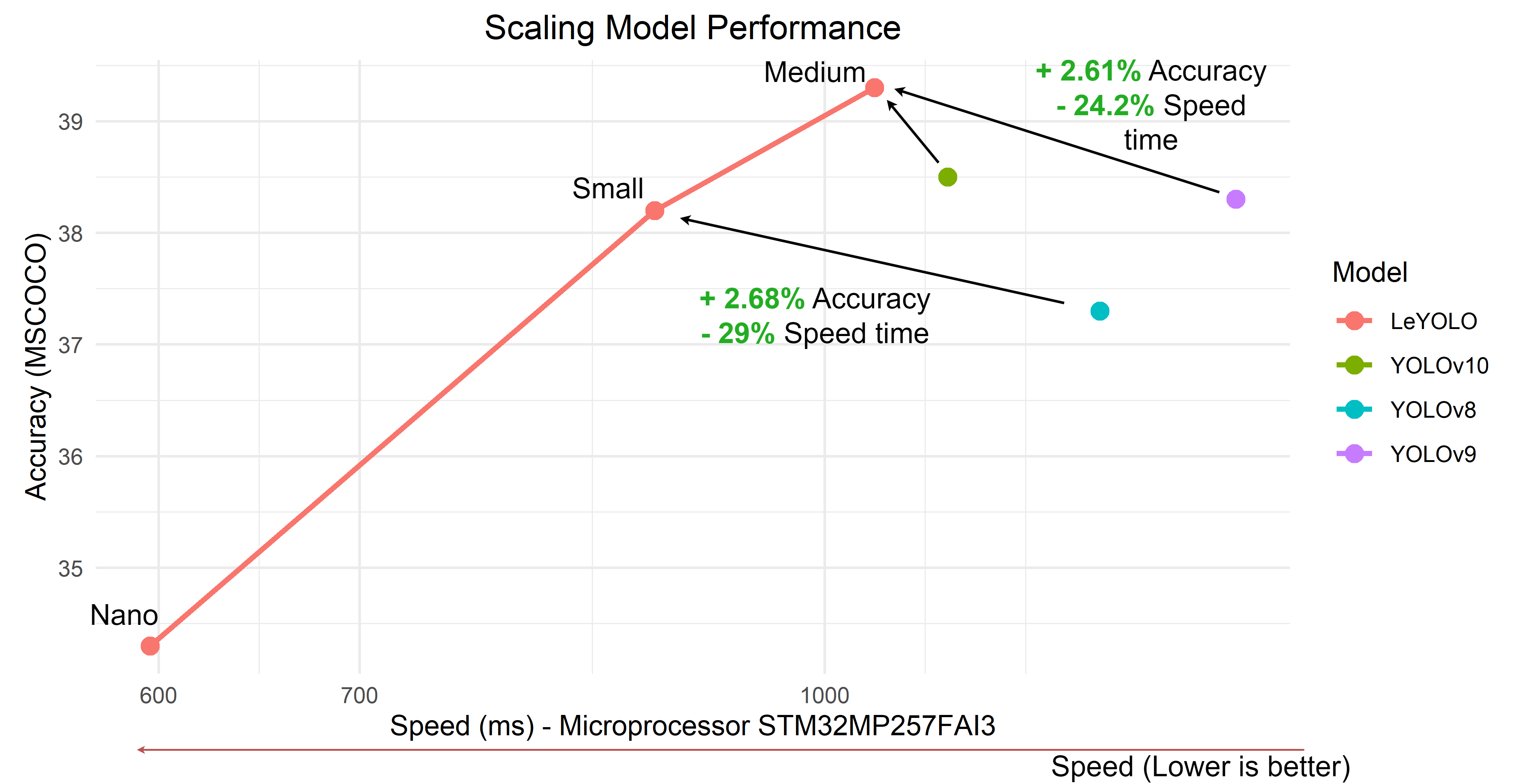}
    \caption{Comparison between LeYOLO and mainline modern YOLO, showing better precision for lower speed detection on STM32MP257FAI3.}
    \label{fig:yolos}
\end{figure}

This paper introduces two principal contributions. 
\begin{enumerate}
    \item The first is an alternative to SSDLite called \textbf{LeNeck} that bridges the gap between low-parameter classifiers and small-scale YOLO models. Using LeNeck instead of SSDLite, we maintain a similar inference speed while achieving much better accuracy (Figure \ref{fig:ssdlite}).
    \item The second contribution is \textbf{LeYOLO} - a new family of lightweight and efficient YOLO models. LeYOLO matches the precision of smaller YOLO variants while significantly improving inference speed on microcontrollers (Figure \ref{fig:yolos}).  
\end{enumerate}

Our findings show that this approach competes with YOLO models at comparable scales. We demonstrate that optimizing neural network architecture for object detection is possible through a new scaling method between lightweight classifiers and YOLO models.

\section{Related Work}
\label{chap:relatedwork}


Our work focuses on developing an optimal architecture for object detection by combining two key approaches: object detectors optimized for speed and low-cost classifiers designed to minimize parameter count using well-established techniques. LeYOLO and LeNeck incorporate elements known for their efficiency in reducing parameter counts. Specifically, we leverage inverted bottlenecks, first introduced in MobileNetV2 \cite{sandler_mobilenetv2_2018} and later refined by EfficientNet \cite{tan_efficientnet_2019, tan_efficientnetv2_2021} and GhostNet \cite{han_ghostnet_2020, tang_ghostnetv2_2022}. Pointwise \cite{lin_network_2014} and depthwise convolutions are crucial components in architecture optimization, contributing significantly to models like MNASNet \cite{tan_mnasnet_2019}.  

The rise of low-cost classifiers led to SSDLite, an optimized SSD variant incorporating grouped convolutions based on MobileNets. Originally designed to lower detection costs using VGG \cite{simonyan_very_2015}, SSDLite shares similarities with early YOLO models \cite{redmon_yolov3_2018}.  Since then, no method has significantly outperformed SSDLite, though SSDLiteX \cite{ssdlitex} has attempted to enhance its performance.  

On the YOLO side, research has explored parameter reduction in mainline architectures. Efforts from tinier-yolo, efficient yolo, mobile densenet, and others \cite{fang_tinier-yolo_2020, yang_ts-yoloefficient_2022, hajizadeh_mobiledensenet_2023, wang_efficient_2020} have integrated lightweight classifier elements like depthwise convolutions and older techniques such as fire modules \cite{iandola_squeezenet_2016} to minimize parameter usage.  

EfficentDet \cite{wang_efficient_2020} shares our model's central philosophy: using layers with low computational cost (concatenation and additions, depthwise and pointwise convolutions). However, EfficientDet requires too much semantic information and too many blocking states (waiting for previous layers, complex graphs), which makes it difficult to keep up with fast execution speed.

SiSO \cite{chen_you_2021} presents an interesting approach to object detection. The authors of YOLOF opted for a model neck with a single input and output. While this design is theoretically faster and more computationally efficient, the YOLOF paper reveals a significant accuracy drop when comparing a single-output neck (Single-in, Single-out - SiSO) to a multi-output neck (Single-in, Multiple-out - SiMO).

More recently, YOLOX \cite{ge_yolox_2021} and YOLOv9 have introduced lightweight alternatives with reduced parameter counts. YOLOX replaces standard convolutions with depthwise convolutions of larger kernel sizes and processes smaller image inputs. YOLOv9 substantially contributes to parameter and information optimization but focuses on standard YOLO scaling rather than mobile-friendly architectures. 

Finally, Tinyssimo YOLO \cite{moosmann_tinyissimoyolo_2023}, built upon early YOLO models \cite{redmon_you_2016}, focuses on reducing computational costs to enable object detection on microcontrollers operating within the milliwatt power range. However, it falls short of achieving the accuracy and efficiency of even the smallest YOLO variants or SSDLite-based classifiers.

\section{Optimizing Real-time Object Detectors for Microcontrollers}
\label{chap:block}
Modern object detectors rely on architecture blocks that fully exploit modern hardware. Standard convolutions and parallel or multi-branch structures are commonly used. However, these designs are too resource-intensive for microcontrollers.
Designed to be highly efficient, LeYOLO core building block optimizes both parameters (memory usage) and mAP (accuracy). It builds on a well-known structure called the inverted bottleneck, commonly used in efficient neural networks like MobileNets \cite{howard_mobilenets_2017, sandler_mobilenetv2_2018, howard_searching_2019, mehta_mobilevit_2022} and EfficientNets \cite{wang_efficient_2020, tan_efficientnetv2_2021}.

\textbf{How It Works.} Instead of using large, expensive filters to process images, LeYOLO breaks the process into smaller, more efficient steps using three main convolution layers. Our block applies a $1\times1$ convolution that projects the feature maps channels $C$ from $x \in \mathbb{R}^{B,C,H,W}$  into a d-dimensional tensor (where $d \geq C$). After that, an $k\times k$ depthwise convolution processes spatial features efficiently. Finally, another $1\times 1$ pointwise convolution brings the channels back to the original size.
While many papers utilizing inverted bottlenecks modify the final pointwise convolution to output a different number of channels than the input, LeNeck and LeYOLO do not follow this approach. Instead, we rely solely on separate standard convolutions when transitioning between feature map sizes after downsampling. These convolutions inherently adjust both the number of channels and the feature map size, eliminating the need for additional transformations within the inverted bottleneck.

\textbf{Optimization Trick.} Normally, the first $1 \times 1$ convolution expands the channels before processing. However, if the number of channels does not need to change (if $C == d$), we remove the first pointwise convolution. This small change significantly reduces the number of computations, especially in early layers where images are large.

\textbf{Impact on Speed and Accuracy.} Eliminating unnecessary computations makes the network faster and more efficient while maintaining high accuracy (Section \ref{chap:deepanalysis}). This optimization is especially beneficial for running object detection models on low-power, resource-constrained devices. For comparison, SSDLite only begins sharing semantic information at the P4 level\footnote{P4 means the semantic level of information corresponding to the input size divided by $2^4$.}, whereas classical YOLO and modern object detectors begin at P3, which provides richer spatial details but at a higher computational cost. By strategically reducing redundant computations in the early layers, LeNeck achieves the same speed as SSDLite while leveraging the more informative P3 level. This results in improved detection performance without added computational overhead.
The model uses the SiLU activation function ($\sigma$), like in modern YOLO versions (YOLOv7, YOLOv9) for improved performance.
We define the input and output dimensions as C and the expanded dimension as d. For filters \(W_{1} \in \mathbb{R}^{1,1,C, d}\), \(W_{2} \in \mathbb{R}^{k,k,1,d}\), and \(W_{3} \in \mathbb{R}^{1,1,d, C}\), our approach can be represented as follows:
\begin{equation}
    y = 
\begin{cases}
  W_{3} \otimes \sigma[W_{2} \otimes \sigma(W_{1} \otimes x)]& \text{if $d \neq C$} \\
  W_{3} \otimes \sigma[W_{2} \otimes \sigma(W_{1} \otimes x)] & \text{if $d = C$ and $W_{1}=$ True} \\
  W_{3} \otimes \sigma[W_{2} \otimes (x)] & \text{if $d = C$ and $W_{1}=$ False} \\
\end{cases}
\end{equation}

\subsection{LeYOLO Backbone}

Our implementation involves minimizing inter-layer information exchange in the form of \(I(X;h_{1}) \geq I(X;h_{2}) \geq ... \geq I(X;h_{n})\), with \(n\) equal to the last hidden layer of the neural network \textbf{backbone}, by ensuring that the number of input/output channels never exceeds a difference ratio of 6 from the first hidden layer through to the last.
Also, rather than augmenting computational complexity in our model like \cite{wang_yolov7_2023, wang_yolov9_2024, hinton_how_2023, cai_reversible_2023}, we opted to scale it more efficiently, integrating Han's et al. \cite{han_rethinking_2021} inverted bottleneck theory which stated that pointwise convolutions should not overpass a ratio of 6 in inverted bottleneck. 

\begin{table}[ht]
    \centering
    \small
    \caption{LeYOLO backbone neural network architecture.}
     \begin{tabular}{|cccccc|}
        \hline
        Input & Operator & exp size & out size & NL & s \\
        \hline
        P0 & conv2d, 3x3 & - & 16 & SI & 2 \\
        P1 & conv2d, 1x1 & 16 & 16 & SI & 1 \\
        P1 & bneck, 3x3, \textbf{pw=False} & 16 & 16 & SI & 2 \\
        
        P2 & bneck, 3x3 & \textbf{96} & 32 & SI & 2 \\
        P3 & bneck, 3x3 & 96 & 32 & SI & 1 \\
        
        P3 & bneck, 5x5 & 96 & 64 & SI & 2 \\
        P4 & bneck, 5x5 & 192 & 64 & SI & 1 \\
        P4 & bneck, 5x5 & 192 & 64 & SI & 1 \\
        P4 & bneck, 5x5 & 192 & 64 & SI & 1 \\
        P4 & bneck, 5x5 & 192 & 64 & SI & 1 \\

        P4 & bneck, 5x5 & \textbf{576} & 96 & SI & 2 \\
        P5 & bneck, 5x5 & 576 & 96 & SI & 1 \\
        P5 & bneck, 5x5 & 576 & 96 & SI & 1 \\
        P5 & bneck, 5x5 & 576 & 96 & SI & 1 \\
        P5 & bneck, 5x5 & 576 & 96 & SI & 1 \\
        \hline
    \end{tabular}
    \label{tab:my_label}
\end{table}

\subsection{LeNeck - General-Purpose Object Detector}

\begin{figure}
    \centering
    \includegraphics[width=1\linewidth]{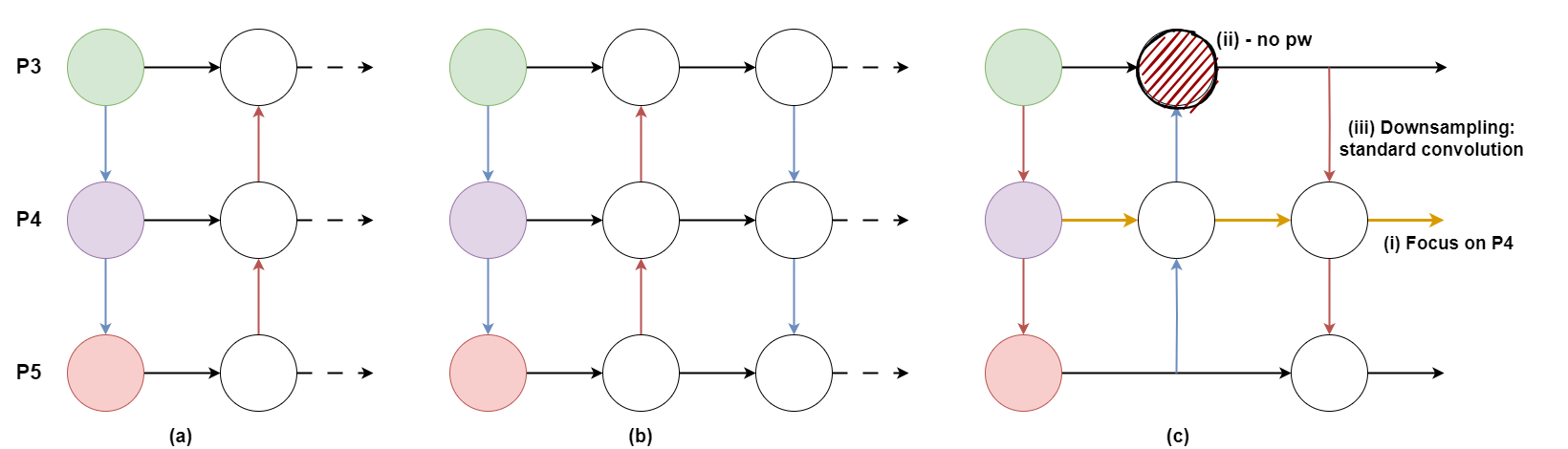}
    \caption{Difference between proposed LeYOLO neck as an efficient semantic feature aggregator. (a) Correspond to FPN \cite{ghiasi_nas-fpn_2019}. (b) Represent PANnet \cite{zhao_pyramid_2017}. Finally, (c) is our proposed solution.}
    \label{fig:neck}
\end{figure}

\textbf{Neck.} In object detection, we call the neck the part of the model that aggregates several levels of semantic information, sharing extraction levels from more distant layers to the first layers. Historically, researchers have used a PANet \cite{zhao_pyramid_2017} or FPN \cite{lin_feature_2017} to share feature maps efficiently, enabling multiple detection levels by linking several semantic information \(P_i\) to the PANet and their respective outputs as depicted in Figure \ref{fig:neck}.(a).\newline
To create LeNeck, we have identified a very important aspect in the composition of deep neural networks.
We noticed that there is consistently a significant repetition of layers at the semantic level equivalent to P4. We found this in all MobileNets \cite{howard_mobilenets_2017, sandler_mobilenetv2_2018, howard_searching_2019}, in the optimization of inverted bottlenecks in EfficientNets \cite{tan_efficientnet_2019, tan_efficientnetv2_2021} and EfficientDet \cite{tan_efficientdet_2020}, as well as in more recent architectures with self-attention mechanisms like MobileViTs \cite{mehta_mobilevit_2022, mehta_separable_2022, wadekar_mobilevitv3_2022}, EdgeNext \cite{maaz_edgenext_2023}, and FastViT \cite{vasu_fastvit_2023}, which are designed for speed. Even more interestingly, models designed by Neural Architecture Search (NAS) \cite{tan_mnasnet_2019, howard_searching_2019, tan_efficientnet_2019} also utilize this pattern.
Therefore, we introduce LeNeck, an efficient semantic feature aggregator that utilizes the P4 semantic level as the primary conduit for merging information from P3 and P5 (Figure \ref{fig:neck}.(i)). Computation at P3 and P5 is performed only once, ensuring efficiency (P3 uses too much spatial size, and P5 uses a very expanded number of channels).

We reduce the computation, especially at P3 level, because of the large spatial size, by removing the first pointwise convolution (Figure \ref{fig:neck}.(ii)). After an ablation study (Section \ref{chap:ablation}) performed on the LeYOLO nano-scaled backbone, we took the opportunity to remove time-costly pointwise convolutions since the input channels from the backbone P3 concatenated with the upsampled features from P4 results in the d-dimension required by the in-between depthwise convolution from our optimized inverted bottleneck presented in section \ref{chap:block}.
The number of input channels, as well as the number of expanded channels from the inverted bottleneck, never exceeds 6. Input from P3 is \(32 C\) while the very last hidden layer of the LeYOLOs neck expanded channels \(d\) equals \(192\).

We improve the accuracy by paying careful attention to stride details. As standard convolutions are not very parameter-efficient and computationally friendly, we thought of a way, in line with our low number of channels, and in regards to computation with strided standard convolutions, to use them two times. From \(P3\) to \(P4\), and from \(P4\) to \(P5\) (Figure \ref{fig:neck}.(iii)). 

\subsection{Ablation Study.}
\label{chap:ablation}

An ablation study in machine learning is a research method that tests the impact of specific layers, features, or techniques by disabling or replacing them. The goal is to identify which components are crucial for the model's performance, guiding the development of a fully optimized object detector. We use LeYOLO in its entirety (backbone + LeNeck) for the ablation study to refine both contributions (Table \ref{tab:ablation}).

We first explored various kernel size configurations. While larger kernels generally improve performance, they also demand more computing resources. The optimal choice was a $5 \times 5$ convolution after P4 downsampling. 

Following insights from ConvNeXt, we employed separate convolutions for downsampling. However, using a $3 \times 3$ kernel instead of $5 \times 5$ in this setup led to better results.

Finally, we made two critical optimizations: reducing the expansion ratio in the inverted bottleneck from 3 to 2 and eliminating the costly first pointwise convolution in the early layers of the backbone and at the P3 level within the neck. These modifications significantly reduced the computational cost of the model while incurring only a minor precision loss of -0.3 mAP, which we deemed negligible given the efficiency gains.

\begin{table}[ht]
    \centering
    \caption{Neck and backbone improvement (best of Ablation study).}
    \begin{tabular}{|c|c|c|}
        \hline
        Improvements & mAP & GFLOP \\
         \hline
         base (LeYOLO nano) & 34.3 & 2.64  \\
         \hline
         +3x3 only & 32.9 & 2.877\\
         +5x5 only & 34.9 & 3.946\\
         +5x5 after P4 & 34.2 & 3.19 \\
         +Downsampling 3x3 only & 34.6 & 3.011\\
         +no pw backbone and neck & 34.1 & 2.823\\
         +LeNeck expansion ratio of 2 instead of 3 & 34.3 & 2.64\\
         \hline
    \end{tabular}
    \label{tab:ablation}
\end{table}

\begin{figure*}
    \centering
    \includegraphics[width=0.9\linewidth]{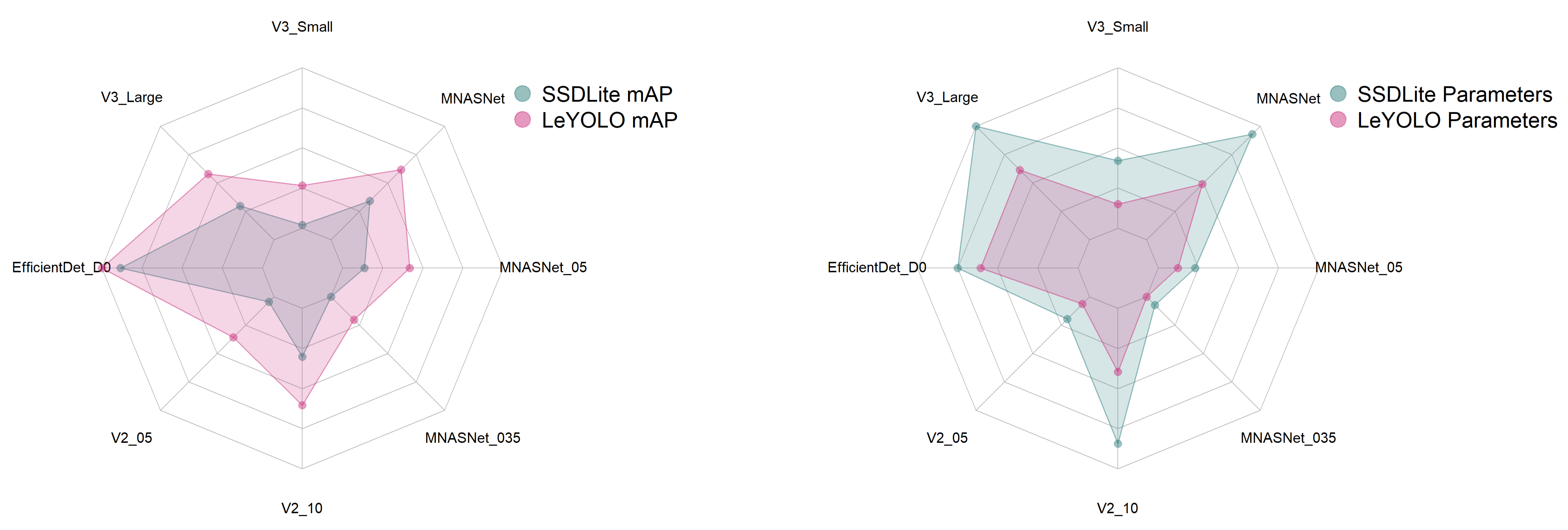}
    \caption{LeNeck compared to SSDLite, with better parameters and precision efficiency}
    \label{fig:radio_leyolo_ssdlite}
\end{figure*}

\section{Experimental results}
\label{chap:exp}
We train each neural network on MSCOCO with the same hyperparameters and data augmentation techniques, such as SGD, with a learning rate of 0.01 and momentum of 0.9. We mostly rely on mosaic data augmentation as well as hsv of \( \{0.015, 0.7, 0.4\}\) and an image translation of \(0.1\). As for the training specificities, we used a 96-batch size over 4 P100 GPUs.
Performance is evaluated on the MSCOCO validation set using mean average precision (mAP). \newline
For LeYOLO, we offer a variety of models inspired by the architectural base presented above. A classic approach involves scaling the number of channels, layers, and input image size. Traditionally, scaling emphasizes channel and layer configurations, sometimes incorporating various scaling patterns (Table \ref{tab:basetrain}). \newline
\begin{table}[ht]
    \centering
    \caption{LeYOLO base training scaling architecture with their respective results}
    \begin{tabular}{|c|cccc|}
        \hline
        Models & Nano & Small & Medium & Large\\
        \hline
        Input spatial size & 640 & 640 & 640 & 768 \\
        Channels ratio & x1 & x1.33 & x1.33  & x1.33 \\
        Layer ratio & x1& x1& x1.33& x1.33\\
        \hline
        mAP&34.3&38.2&39.3&41\\
        \hline
    \end{tabular}
    \label{tab:basetrain}
\end{table}
\newpage
LeYOLO scale from Nano to Large version with scaling related to what EfficientDet brought: \(channels\) from \(1.0\) to \(1.33\), \(layers\) from \(1.0\) to \(1.33\), and spatial size for training purpose from \(640\times640\) to \(768\times768\). Several spatial sizes are used for evaluation purposes, ranging from \(320\times320\) to \(768\times768\). 
We evaluate LeYOLO at reduced spatial sizes, with all results presented in Table \ref{tab:completesoa}, showing the number of FLOPs. We observe a correlation between this metric and the execution speed on low-computation devices (Table \ref{tab:speed}).

We evaluate model speed on two microprocessors: STM32MP257FAI3 and STM32N6570-DK. Both use Arm Cortex cores, balancing low power consumption and efficient processing. These microcontrollers can achieve real-time inference at 320×320 resolution. At 640×640, computation becomes more demanding, but LeYOLO still processes each inference in under a second, outperforming modern state-of-the-art YOLO models (Table \ref{tab:speed}).

\begin{table}[ht]
    \centering
    \small
    \caption{LeYOLO (640x640) inference speed (ms - lower is better) and accuracy ratio on an embedded device (Onnx - STM32MP257FAI3). }
    \begin{tabular}{|l|c|c|}
    \hline
    Models & mAP & Speed(ms) \\
    \hline
    LeYOLO Nano &  \textbf{34.3} & \textbf{596}\\
    LeYOLO Small &  \textbf{38.2} & \textbf{877.9} \\
    \hline
    LeYOLO Medium &  \textbf{39.3} & \textbf{1039}\\
    \hline
    YOLOv10 Nano \cite{wang2024yolov10} &  38.5 & 1099\\
    YOLOv8 Nano \cite{jocher_ultralytics_2023} &  37.3 & 1235 \\
    YOLOv9 Tiny \cite{wang_yolov9_2024} & 38.3 & 1371 \\
    \hline
    \end{tabular}
    \label{tab:speed}
\end{table}

\begin{table}[ht]
    \centering
    \small
    \caption{LeNeck (320x320) inference speed (ms - lower is better) and accuracy ratio on an embedded device (Onnx - STM32MP257FAI3). }
    \begin{tabular}{|c|c|c|c|c|}
    \hline
        Models & SSDLite & \textbf{LeNeck} & SSDLite & \textbf{LeNeck}\\
        \hline
         & \multicolumn{2}{c|}{Speed(ms)} & \multicolumn{2}{c|}{mAP.95}\\
         \hline
        V3-Small&\textbf{146.2}&165.9&16.0&\textbf{21.3}\\        
        V3-Large&\textbf{286.5}&292.3&22&\textbf{28.1}\\
        V2-1.0&259.2&\textbf{256.3}&22.1&\textbf{28.6}\\
        MNASNet 0.5&167.7&\textbf{155.3}&18.5&\textbf{24.6}\\
        MNASNet&306.2&\textbf{262.3}&23&\textbf{28.9}\\
        \hline
        LeYOLO Nano&&165.4&&25.2\\
        \hline
    \end{tabular}
     
    \label{tab:lightweight_leyolo}
\end{table}

\begin{table}[ht]
    \centering
    \small
    \caption{LeNeck (320x320) inference speed (ms - lower is better) and accuracy ratio on an embedded device (Onnx - STM32N6570-DK). }
    \begin{tabular}{|c|c|c|c|c|}
    \hline
        Models & SSDLite & \textbf{LeNeck} & SSDLite & \textbf{LeNeck}\\
        \hline
         & \multicolumn{2}{c|}{Speed(ms)} & \multicolumn{2}{c|}{mAP.95}\\
         \hline
        V3-Small&\textbf{46.91}&50.19&16.0&\textbf{21.3}\\        
        V3-Large&\textbf{121.6}&129.2&22&\textbf{28.1}\\
        V2-1.0&104&\textbf{103.5}&22.1&\textbf{28.6}\\
        MNASNet 0.5&86.31&\textbf{36.03}&18.5&\textbf{24.6}\\
        MNASNet&161.3&\textbf{53.78}&23&\textbf{28.9}\\
        \hline

    \end{tabular}
     
    \label{tab:lightweight_leyolo_other}
\end{table}


\begin{table}[ht]
    \centering
    \small
     \caption{LeNeck performance compared to lightweight classifier on MSCOCO object detection downstream tasks with SSDLite}
    \begin{tabular}{|c|c|c|c|c|}
    \hline
        Models & SSDLite & \textbf{LeNeck} & SSDLite & \textbf{LeNeck}\\
        \hline
         & \multicolumn{2}{c|}{Parameters(M)} & \multicolumn{2}{c|}{mAP.95}\\
         \hline
        V3-Small&2.49&\textbf{1.34}&16.0&\textbf{21.3}\\
        V3-Large&4.97&\textbf{3.33}&22&\textbf{28.1}\\
        EfficientDetD0&3.9&\textbf{3.29}&34.6&\textbf{37.1}\\
        V2-0.5&1.54&\textbf{0.98}&16.6&\textbf{23.3}\\
        V2-1.0&4.3&\textbf{2.39}&22.1&\textbf{28.6}\\
        MNASNet 0.35&1.02&\textbf{0.7}&15.6&\textbf{20.0}\\
        MNASNet 0.5&1.68&\textbf{1.22}&18.5&\textbf{24.6}\\
        MNASNet&4.68&\textbf{2.8}&23&\textbf{28.9}\\
        \hline

    \end{tabular}
   
    \label{tab:lightweight_leneck}
\end{table}

\textbf{SOTA Backbone Results with LeNeck.} We integrate several sota low-parameter backbones with \textbf{LeNeck}. No matter the backbone used, all channel numbers, \(P3, P4\) and \(P5\) repetition specificities stay the same.
At \(P3\), the first pointwise convolution is never used, like in baseline LeYOLO, resulting in the first filter being the depthwise convolution of the exact size of the backbone equivalent input number of channels. 
\newline
LeNeck, as a general object detector for lightweight classifiers, keeps the same number of channels\footnote{\(32\) to \(96\) channels with an extension ratio of \(2\)} and layer repetition\footnote{repetition \(l=3\) } as the LeYOLO Nano version.
From a variety of lightweight classifiers with a low number of parameters and FLOP, LeNeck outperformed SSDLite in every aspect of what we expect from a low-cost model - better parameter scaling, better precision, and finally, better inference speed\footnote{From the variety of available, fully reproducible and testable model on STM32 devices} - with inference speed results described in tables \ref{tab:lightweight_leyolo} - \ref{tab:lightweight_leyolo_other} and parameters-to-accuracy efficiency described in Table \ref{tab:lightweight_leneck} and Figure \ref{fig:radio_leyolo_ssdlite}.
\newpage
\textbf{Low-end Microcontrollers.} Beyond real-time processing, we further benchmark LeYOLO against modern YOLO models on various low-end microcontrollers, as shown in Table \ref{tab:lowendspeed}, using the LeYOLO-Small variant (YOLOv10 is not compatible with these kinds of microcontrollers). According to Table \ref{tab:completesoa}, LeYOLO-Small matches YOLOv8 to YOLOv10 in accuracy. Additionally, it proves to be more inference-efficient on these microcontrollers, extending YOLO's capability to run effectively on low-end devices.

\begin{table}[ht]
    \centering
    \small
    \caption{LeYOLO (640x640) inference speed on embedded devices. }
    \hspace*{-0.6cm}\begin{tabular}{|l|c|c|c|}
    \hline
    Device & LeYOLO Small & YOLOv8 & YOLOv9\\
    \hline
    & \multicolumn{3}{c|}{Speed (s)} \\
    \hline
    STM32H74l-DISCO&\textbf{12.3}&13.7&13.6\\
    STM32F769l-DISCO&\textbf{19}&21.5&22.1\\
    STM32F746G-DISCO&\textbf{20}&25&24.5\\
    STM32F469I-DISCO&\textbf{54.6}&73.6&72.5\\
    \hline
    \end{tabular}
    \label{tab:lowendspeed}
\end{table}


\subsection{Deeper Analysis}
\label{chap:deepanalysis}

\begin{table}[ht]
    \centering
    \small
    \caption{LeYOLO Small (640x640) inference speed improvements}
    \hspace*{-0.6cm}\begin{tabular}{|l|c|c|c|}
    \hline
    Device & LeYOLO Small & pw & exp x3\\
    \hline
    & \multicolumn{3}{c|}{Speed (s)} \\
    \hline
    STM32H74l-DISCO&\textbf{12.37}&13.52&14.87\\
    STM32F769l-DISCO&\textbf{19.04}&20.64&22.68\\
    STM32F746G-DISCO&\textbf{20}&22.36&24.73\\
    STM32F469I-DISCO&\textbf{54.6}&59.23&65.28\\
    \hline
    \end{tabular}
    \label{tab:lowendspeedbchmark}
\end{table}

To further analyze the speed results, we highlight the purpose of our inverted bottleneck with an optional pointwise layer (see Section \ref{chap:block} for more information). In LeYOLO Small, pointwise convolution at high spatial resolutions in the backbone and at the P3 level in LeNeck leads to a minimal accuracy improvement, as shown in Section \ref{chap:ablation}. Compared to the classical Inverted Bottleneck, our solution saves 8.5\% of inference speed on all STM32 benchmarked in the paper (Table \ref{tab:lowendspeedbchmark}-pw).
Unlike standard YOLO architectures, which rely on deep layer repetition with fewer channels, LeYOLO achieves better efficiency by using an expansion ratio of 2 instead of 3 while maintaining a minimum depth repetition of 3. This design choice improves inference speed by 17\% while preserving accuracy, as confirmed by our ablation study (Table \ref{tab:lowendspeedbchmark}-exp x3).

Our model is nearly as fast as SSDLite while achieving significantly better accuracy. The slight speed difference comes from feature map size—SSDLite starts at P4, while we begin at P3. However, LeNeck remains lightweight enough to compete with SSDLite in speed. Since it retains richer spatial information, LeNeck also performs better at detecting various object sizes (Table \ref{tab:completesoa}).

\subsection{More Datasets Analysis}

\begin{table}[htbp]
\centering
\caption{Comparison of LeYOLO nano with sota YOLO models (mAP$_{50}$)}
\begin{tabular}{|l|c|c|c|c|c|}
\hline
\textbf{Models} & \textbf{Params} & \textbf{African} & \textbf{Brain} & \textbf{WGISD} & \textbf{Global} \\
 & (M) & \textbf{WildLife} & \textbf{Tumor} &  &\textbf{Wheat} \\
\hline
LeYOLO & 1.0 & 0.870 & 0.500 & 0.694 & 0.956 \\
YOLOv5 & 2.5 & 0.896 & 0.497 & 0.776 & 0.967 \\
YOLOv8 & 3.2 & 0.907 & 0.519 & 0.795 & 0.969 \\
YOLOv9 & 2.0 & 0.905 & 0.508 & 0.791 & 0.970 \\
YOLOv10 & 2.3 & 0.865 & 0.488 & 0.723 & 0.957 \\
\hline
\end{tabular}
\label{tab:yolo_comparison_alldataset}
\end{table}

\begin{figure}[ht]
    \centering
    \includegraphics[width=1.0\linewidth]{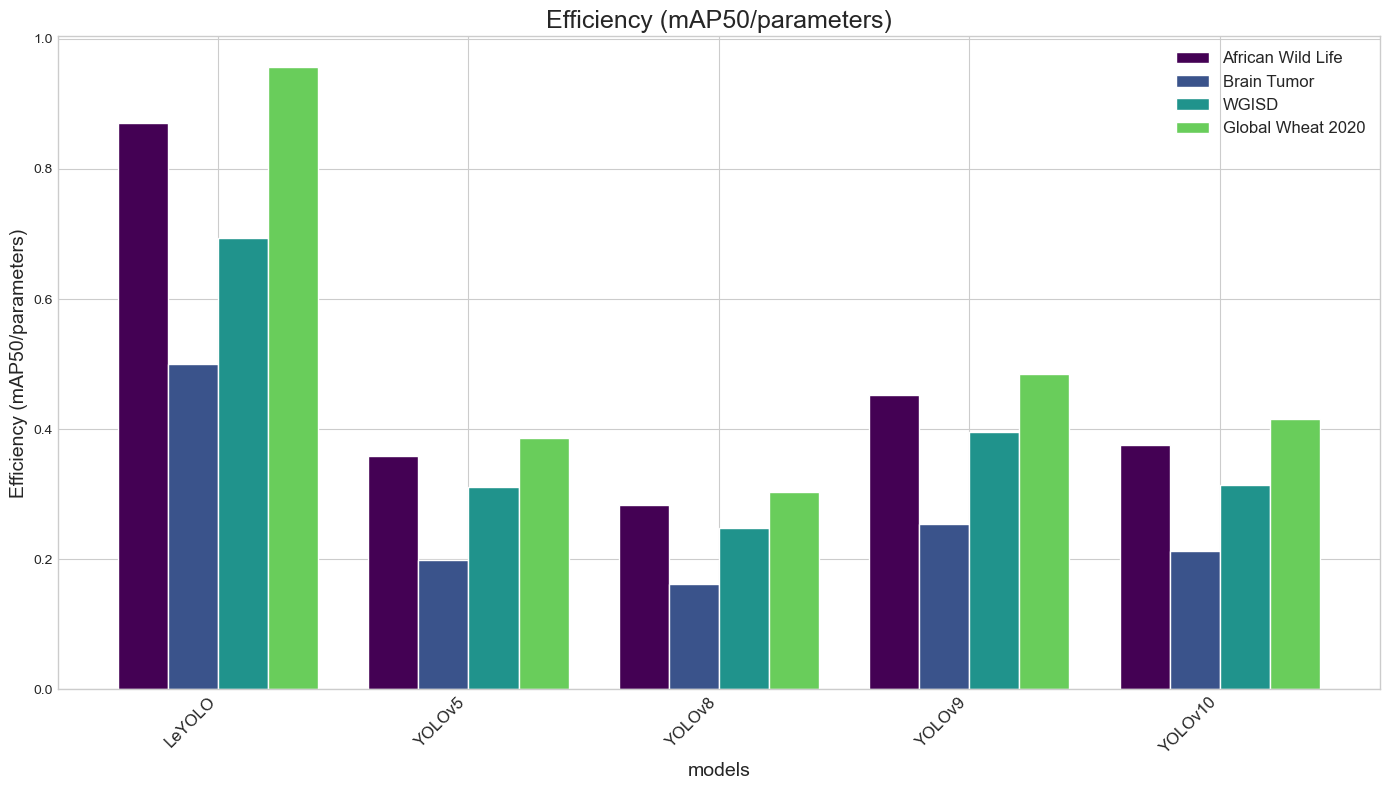}
    \caption{Efficiency Comparison: Accuracy-to-Parameter Ratios for Mainline YOLO on Real-World Datasets}
    \label{fig:acc_param_alldatasets}
\end{figure}

To further demonstrate LeYOLO’s capabilities compared to standard YOLO models in terms of accuracy, we evaluated both on real-world datasets with fewer samples and classes. All models were initially trained on the MSCOCO dataset and fine-tuned on four target datasets: African Wildlife, Brain Tumor, WGISD, and Global Wheat 2020. Overall, the accuracy levels across all models are relatively similar. However, what stands out is the precision-to-parameter ratio of each model. LeYOLO consistently outperforms the standard YOLO variants across all four datasets, achieving significantly higher accuracy relative to the number of parameters in the network (Figure \ref{fig:acc_param_alldatasets} and Table \ref{tab:yolo_comparison_alldataset}).

\begin{table*}
    \centering
    \caption{State-of-the-art lightweight object detector compatible with STM32 microcontrollers.}
    \small
    \resizebox{2.0\columnwidth}{!}{%
    \begin{tabular}{lllcccccrr}
    \hline
        Models & Input Size & mAP & mAP50 & mAP75 & S & M & L & FLOP(G)&Parameters (M)\\
    \hline
         MobileNetv3-S\cite{howard_mobilenets_2017}&320& 16.1 &  -&-&-&-&-&\textbf{0.32}&1.77\\
         MobileNetv2-x0.5\cite{sandler_mobilenetv2_2018}&320 &16.6   &- &-&-&-&-&0.54&1.54\\
         MnasNet-x0.5\cite{tan_mnasnet_2019}&320&18.5 &-&-&-&-&-&0.58&1.68\\
         \textbf{LeYOLO-Nano}&320&\textbf{25.2}&\textbf{37.7}&\textbf{26.4}&\textbf{5.5}&\textbf{23.7}&\textbf{48.0}&0.66&\textbf{1.1}\\
        \hline
        MobileNetv3\cite{howard_searching_2019}&320&22 &-&-&-&-&-&1.02&3.22\\
        \textbf{LeYOLO-Small}&320&\textbf{29}&42.9&30.6&6.5&29.1&53.4&1.126&1.9\\
        \textbf{LeYOLO-Nano}&480&\textbf{31.3}&\textbf{46}&\textbf{33.2}&\textbf{10.5}&\textbf{33.1}&\textbf{52.7}&1.47&1.1\\
        MobileNetv2\cite{sandler_mobilenetv2_2018}&320&22.1 &-&-&-&-&-&1.6&4.3\\
        MnasNet\cite{tan_mnasnet_2019}&320&23&-&-&-&-&-&1.68&4.8\\
        \hline
        \textbf{LeYOLO-Small}&480&\textbf{35.2}&50.5&37.5&13.3&38.1&55.7&\textbf{2.53}&\textbf{1.9}\\
        MobileNetv1\cite{howard_mobilenets_2017}&320&22.2  &-&-&-&-&-&2.6&5.1\\
        \textbf{LeYOLO-Medium}&480&\textbf{36.4}&\textbf{52.0}&\textbf{38.9}&\textbf{14.3}&\textbf{40.1}&\textbf{58.1}&3.27&2.4\\
        \hline
        \textbf{LeYOLO-Small}&640&\textbf{38.2}&\textbf{54.1}&\textbf{41.3}&\textbf{17.6}&\textbf{42.2}&\textbf{55.1}&\textbf{4.5}&\textbf{1.9}\\
        YOLOv5-n\cite{jocher_ultralyticsyolov5_2022}&640&28&45.7&-&-&-&-&4.5&1.9\\
        EfficientDet-D0\cite{tan_efficientdet_2020}&512&33.80 &52.2&35.8&12&38.3&51.2&5&3.9\\
        \hline
        \textbf{LeYOLO-Medium}&640&\textbf{39.3}&\textbf{55.7}&\textbf{42.5}&\textbf{18.8}&\textbf{44.1}&\textbf{56.1}&\textbf{5.8}&\textbf{2.4}\\
        YOLOv9-Tiny\cite{wang_yolov9_2024}&640&38.3&53.1&41.3&-&-&-&7.7&2\\
        \hline
        \textbf{LeYOLO-Large}&768&\textbf{41}&\textbf{57.9}&\textbf{44.3}&\textbf{21.9}&\textbf{46.1}&\textbf{56.8}&\textbf{8.4}&\textbf{2.4}\\
    \hline
    \end{tabular}%
    }
    \label{tab:completesoa}
\end{table*}

\newpage
\section{Discussions}

\textbf{LeNeck:} Considering the cost-effectiveness of LeNeck, there is a significant opportunity for experimentation across different backbones of state-of-the-art classification models. LeYOLO emerges as a promising alternative to SSD and SSDLite. The promising results achieved on MSCOCO with our solution suggest potential applicability to other classification-oriented models We focused our optimization efforts specifically on \textbf{MSCOCO} and YOLO-oriented networks. However, we encourage experimentation with our solution on other datasets as well.

\textbf{Computational efficiency: }We have implemented a new scaling for YOLO models, proving that it is possible to achieve very high levels of accuracy while using very few computational resources (FLOP). LeYOLO provides very fast results within embedded devices.

\section{Conclusion}
As we try to offer thorough theoretical insights from state-of-the-art neural networks to craft optimized solutions, we acknowledge several areas for potential improvement, and we cannot wait to see further research advancements with \textbf{LeNeck} and \textbf{LeYOLO}.

Throughout this paper, we introduced several key optimizations:
\begin{itemize}
    \item \textbf{Improved classifier object detection task performance:} For a given parameter budget, LeNeck surpasses state-of-the-art low-cost classifiers combined with SSDLite by reducing parameter count while improving MSCOCO accuracy. Integrating LeNeck with existing low-parameter backbones enhances accuracy and efficiency across multiple scales.
    \item \textbf{A viable alternative to tiny-scale YOLO models:} LeYOLO's optimized backbone and LeNeck outperform tiny YOLO variants in object detection. The architectural choices behind LeYOLO's backbone result in superior scaling and a better accuracy-to-parameter and FLOP ratio.
    \item \textbf{Enhanced inference speed:} LeYOLO and LeNeck achieve better inference speed than state-of-the-art low-parameter object detectors, thanks to their optimized architecture.
\end{itemize}

Our contributions are particularly effective on mobile, embedded, and low-power devices, moving closer to an ideal balance between parameter efficiency and detection performance. Reducing model size while maintaining accuracy enables object detection directly on small devices with minimal computational overhead. This step-by-step refinement brings YOLO models closer to practical edge AI applications.

\section*{Acknowledgment}
This work was supported by Chips Joint Undertaking (Chips JU) in EdgeAI “Edge AI Technologies
for Optimised Performance Embedded Processing” project, grant agreement No 101097300.


\newpage
%
\bibliographystyle{IEEEtran} 
\bibliography{IEEEtranBST/bare_conf}

\end{document}